\documentclass[acmsmall]{acmart}

\AtBeginDocument{%
  \providecommand\BibTeX{{%
    \normalfont B\kern-0.5em{\scshape i\kern-0.25em b}\kern-0.8em\TeX}}}

\setcopyright{acmcopyright}
\copyrightyear{2023}
\acmYear{2023}
\acmDOI{XXXXXXX.XXXXXXX}

\acmJournal{TIST}
\acmVolume{XX}
\acmNumber{X}
\acmArticle{XXX}
\acmMonth{7}

\usepackage{amsmath,amsthm,amsfonts}
\usepackage{color}
\usepackage{hyperref}

\usepackage{amsmath}

\usepackage{algorithm}
\usepackage{algorithmic}
\usepackage{graphicx}
\usepackage{textcomp}
\usepackage{xcolor}
\usepackage{url}
\usepackage{subcaption}
\usepackage{arydshln}
\usepackage{amsthm}
\usepackage{multirow}
\usepackage{hyperref}
\usepackage{enumitem}

\usepackage{pifont}

\newcommand{\xmark}{\text{\ding{55}}}
\usepackage{microtype}

\begin{document}

\title{DDNAS: Discretized Differentiable Neural Architecture Search for Text Classification}

\author{Kuan-Chun Chen}
\affiliation{%
 \institution{National Cheng Kung University}
 \city{Tainan}
 \country{Taiwan}}
\email{champion516615@gmail.com}

\author{Cheng-Te Li}
\affiliation{%
 \institution{National Cheng Kung University}
 \city{Tainan}
 \country{Taiwan}}
\email{chengte@mail.ncku.edu.tw}

\author{Kuo-Jung Lee}
\affiliation{%
 \institution{National Cheng Kung University}
 \city{Tainan}
 \country{Taiwan}}
\email{kuojunglee@ncku.edu.tw}

\renewcommand{\shortauthors}{K.-C. Chen, et al.}

\begin{abstract} 
Neural Architecture Search (NAS) has shown promising capability in learning text representation. However, existing text-based NAS neither performs a learnable fusion of neural operations to optimize the architecture, nor encodes the latent hierarchical categorization behind text input. This paper presents a novel NAS method, Discretized Differentiable Neural Architecture Search (DDNAS), for text representation learning and classification. With the continuous relaxation of architecture representation, DDNAS can use gradient descent to optimize the search. We also propose a novel discretization layer via mutual information maximization, which is imposed on every search node to model the latent hierarchical categorization in text representation. Extensive experiments conducted on eight diverse real datasets exhibit that DDNAS can consistently outperform the state-of-the-art NAS methods. While DDNAS relies on only three basic operations, i.e., convolution, pooling, and none, to be the candidates of NAS building blocks, its promising performance is noticeable and extensible to obtain further improvement by adding more different operations. 
\end{abstract}



\maketitle

\section{Introduction}
\label{sec:introduction}
Text representation and classification is one of the fundamental tasks in natural language processing. A variety of applications can be benefited from text classification, such as sentiment analysis, topic labeling, spam detection, and claim verification. Deep learning techniques, such as convolutional and recurrent neural networks, are widely applied to text representation learning. Recent advances in the Transformer architecture~\cite{trm17}, such as BERT~\cite{bert} and GPT~\cite{gpt2}, further exhibit promising performance on different downstream tasks of natural language understanding. Nevertheless, we still know limited about the optimal model architecture for text representation learning~\cite{nlpsuv1,nlpsuv2}.

Techniques of \textit{Neural Architecture Search} (NAS) are developed to automatically learn entire neural models or individual neural cell architectures. Typical NAS algorithms, such as ENAS~\cite{enas18}, SMASH~\cite{smash18}, and AmoebaNet-A~\cite{darts19}, have achieved state-of-the-art or competitive performance on various tasks in computer vision and natural language inference. Although promising results can be produced by intelligently customizing the architecture design, most of these methods belong to learning the \textit{micro search space}, in which the controller designs modules or building blocks. Existing studies~\cite{enas18,textnas20} had shown that the \textit{macro space search} that allows the controller to design the entire network model is more effective for text classification than the micro space search. TextNAS~\cite{textnas20} is the state-of-the-art approach to learning the neural architecture within the macro search space, which is further designed for text representation learning.

We think the macro-space neural architecture search for text representation and classification, specifically TextNAS~\cite{textnas20}, can be improved in two aspects. First, TextNAS performs searching over a countable set of candidate architectures. Such an approach is difficult to capture and exploit the \textit{learnable fusion} of various neural operations, e.g., convolution and pooling, to generate the intermediate feature representations. That said, the macro search space of TextNAS is discrete and non-differentiable. The continuous relaxation of the neural architecture representation, e.g., DARTS~\cite{darts19}, can allow the gradient descent process to optimize the architecture. Second, regarding the learning of text representations, encoding the latent hierarchical categorization for documents in the corpus had brought some improvement for classification performance~\cite{hier1,hier2}. Nevertheless, it is challenging to learn the latent categories of the given text in a hierarchical manner without side information as the model input. To the best of our knowledge, existing NAS-based text classification methods including TextNAS do not take the modeling of hierarchical categorization into account. In this work, without utilizing prior knowledge about text categories as the model input, we aim at bringing hierarchical categorization into neural architecture search. That said, we will encode the latent hierarchical categorization into text representations through the proposed NAS technique.

To tackle these two issues in text representation and classification, in this work, we propose a novel NAS-based method, \textit{\underline{\textbf{D}}iscretized \underline{\textbf{D}}ifferentiable \underline{\textbf{N}}eural \underline{\textbf{A}}rchitecture \underline{\textbf{S}}earch} (\textbf{DDNAS})~\footnote{Code is available at this link: \url{https://github.com/ddnas/ddnas}}. DDNAS relies on DARTS~\cite{darts19} to perform the macro space search. We relax the search space, depicted by a directed acyclic graph, to be continuous so that the neural architecture can be optimized via gradient descent. Most importantly, we present a novel \textit{discretization} layer to tailor the search space for text classification. We discretize the continuous text representation to discrete random variables with multiple states. The discretization layer is imposed into every search node of latent representation. Through information passing from low- to high-level discretization in the intermediate embeddings, DDNAS can encode the latent hierarchical categorization into the final representation of input text, which will be demonstrated via visualization in the experiment. We propose a joint learning mechanism to simultaneously learn the latent categorization in the discretization layer and classify the given text.

\begin{table*}[!t]
\centering
\caption{Comparison of relevant studies on recent NAS for text representation and classification. The meanings of each column in the table: Data Type = ``Targeted Data Type'', Space Type = ``Type of Search Space'', Differentiable = ``Whether the search space is continuous?'' Hier-Categorization = ``Whether it captures latent hierarchical categorization in the data?'', Transformer = `Whether it adopts transformer~\cite{trm17} as the building block?'' Highway = ``Whether it requires highway connections~\cite{highwaynet}?''}
\label{tab:relatedtab}
\resizebox{1.0\textwidth}{!}{%
\begin{tabular}{l|c|c|c|c|c|c}
\hline
 & Data Type & Space Type & Differentiable & Hier-Categorization & Transformer & Highway \\ \hline
AmoebaNet-A~\cite{eanas19} & Image & Micro &  &  &  &  \\ \hline
SMASH~\cite{smash18} & Image & Micro &  &  &  &  \\ \hline
ENAS~\cite{enas18} & Image & Micro &  &  &  & \checkmark \\ \hline
DARTS~\cite{darts19} & Text \& Image & Macro & \checkmark &  &  & \checkmark \\ \hline
TextNAS~\cite{textnas20} & Text & Macro &  &  & \checkmark &  \\ \hline
\textbf{DDNAS (This work)} & \textbf{Text} & \textbf{Macro} & \textbf{\checkmark} & \textbf{\checkmark} & \textbf{} & \textbf{} \\ \hline
\end{tabular}%
}
\end{table*}

We summarize the contributions of this work as follows. 
\begin{itemize}
\item We develop a novel NAS framework, \textit{Discretized Differentiable Neural Architecture Search} (DDNAS), to search for more effective differentiable architectures in the macro space and learn better feature representations for text classification. 
\item A novel discretization layer based on mutual information maximization is proposed to capture the latent hierarchical categorization in text representations without relying on prior knowledge or side information. 
\item Extensive experiments conducted on eight diverse benchmark datasets with different numbers of classes exhibit the promising performance of DDNAS, compared to the state-of-the-art NAS models. The results also show that DDNAS is an effective but compact model architecture learned from a limited number of training samples. We also visualize the discretization outcomes, which show that different classes have various distributions of states.
\item With only three basic operations as the NAS building blocks (convolution, pooling, and none), i.e., using no recurrent and attention components, DDNAS consistently leads to the best performance, which demonstrates the potential of DDNAS.
\end{itemize}

This paper is organized as follows. We first review relevant studies in Section~\ref{sec-related}. Then we present the technical details of our proposed DDNAS in Section~\ref{sec-method}. Experimental evaluation is reported in Section~\ref{sec-exp}. We discuss the issues highlighted by our DDNAS framework in Section~\ref{sec-discuss}, and conclude this work in Section~\ref{sec-conclude}.

\section{Related Work}
\label{sec-related}
\textbf{Neural Architecture Search.}
Neural architecture search (NAS) is the key technique for automatic machine learning, and its goal is to find the best model architecture based on a specified search space. NAS has exhibited excellent performance for tasks in the domains of computer vision~\cite{rlnas17,enas18,darts19} and nature language processing~\cite{adabert20,textnas20}. 
Before the automatic architecture search of recurrent neural networks (RNN), to find better model architectures, Greff et al.~\cite{nas17} investigate eight variants of LSTM~\cite{lstm97} on several supervised tasks, but still concludes the vanilla LSTM mostly outperforms the modifications. Jozefowicz et al.~\cite{nas15} further examine over ten thousand variants of GRU~\cite{gru15} by evolutionary approaches, but cannot find any that would consistently outperform LSTM. 
In recent NAS studies, Zoph and Le~\cite{rlnas17} exploit a recurrent network with reinforcement learning to compose neural network architectures in the form of variable-length strings. ENAS~\cite{enas18} is an efficient NAS algorithm, in which the training scheme is devised to share learnable parameters among architectures, instead of fully training each structure individually, to achieve efficient search. SMASH~\cite{smash18} trains an auxiliary model to dynamically adjust model weights with variable architectures, rather than fully training candidate models from scratch. AmoebaNet-A~\cite{eanas19} evolves neural architectures of image classifiers through a modified tournament selection evolutionary mechanism. Last, Liu et al.~\cite{darts19} further invent a differentiable neural architecture search algorithm, DARTS, which can learn to search the weighted combinations of model operations in a continuous space. The proposed DDNAS in this work will extend DARTS by imposing discretization layers to better encode the latent hierarchical semantics of input text.

\textbf{NAS for Text Representation Learning.}
Recent advances in text representation learning and classification, which rely on the architecture of Transformer~\cite{trm17}, such as BERT~\cite{bert} and GPT~\cite{gpt2}, have achieved significant success on a variety of downstream tasks. Less attention is paid to investigating how NAS can be utilized to learn better text representations. The architecture variants of RNN are first explored. Dodge et al.~\cite{rnnarch19} perform structure learning with group lasso regularization to search sparsity-aware RNN architectures. Merrill et al.~\cite{rnnhier20} come up with a formal hierarchy of the expressive capacity of RNN architectures.
Differentiable NAS (DARTS)~\cite{darts19} are modified for named entity recognition~\cite{nasner19} and BERT compression~\cite{adabert20}. TextNAS~\cite{textnas20} is the most relevant NAS-based method to our work for text classification. TextNAS develops an architecture search space specialized for text representation. TextNAS is treated as a state-of-the-art method, and will be compared to the proposed DDNAS because both aim at tailoring the search space for text classification.

We create a table to compare our work with existing studies on NAS-based representation learning and classification, as presented in Table~\ref{tab:relatedtab}. Six different aspects are being compared. Based on the table comparison, it can be found that AmoebaNet-A~\cite{eanas19}, SMASH~\cite{smash18}, and ENAS~\cite{enas18} search the architecture in the micro space for image classification. While regarding the text classification and the macro search space, the most relevant NAS models are TextNAS~\cite{textnas20} and DARTS~\cite{darts19}. Nevertheless, there are still some fundamental differences. TextNAS cannot search the architecture in the continuous space, i.e., non-differentiable, and requires the multi-head self attention~\cite{trm17} as the building block to have better classification performance. Although DARTS has a continuous search space as our DDNAS, it requires highway connections~\cite{highwaynet} to facilitate increased network depth and allow information to flow across several layers without attenuation. Our proposed DDNAS relies on neither transformers nor highway connections to maintain a more compact design space, which can to some extent avoid overfitting and bring better generalization for a NAS method. More importantly, DDNAS attempts to capture the latent hierarchical categorization~\cite{hiersuv11} of the data input, which is never considered in all of the existing NAS models, through the proposed discretization layers imposed into every search node.

\textbf{Mutual Information Maximization.}
Mutual information maximization had been adopted to enhance the representation learning in computer vision, and obtained promising performance in various downstream tasks. Mutual Information Neural Estimation (MINE)~\cite{mine18} estimates the mutual information between continuous random variables in high dimension through gradient descent, and applies it to generative models and supervised learning with improved performance. Deep InfoMax~\cite{dmim19} maximizes the mutual information between local and global visual features and imposes the prior on model input so that the statistic knowledge can be encoded into unsupervised representations. To the best of our knowledge, Kong et al.~\cite{mim20} is the first attempt to study how mutual information maximization (MIM) between local and global information can be used for language representation learning, and constructs a self-supervised learning task with MIM to generate text representations.

\begin{figure}[!t]
\centering
\includegraphics*[width=0.65\linewidth]{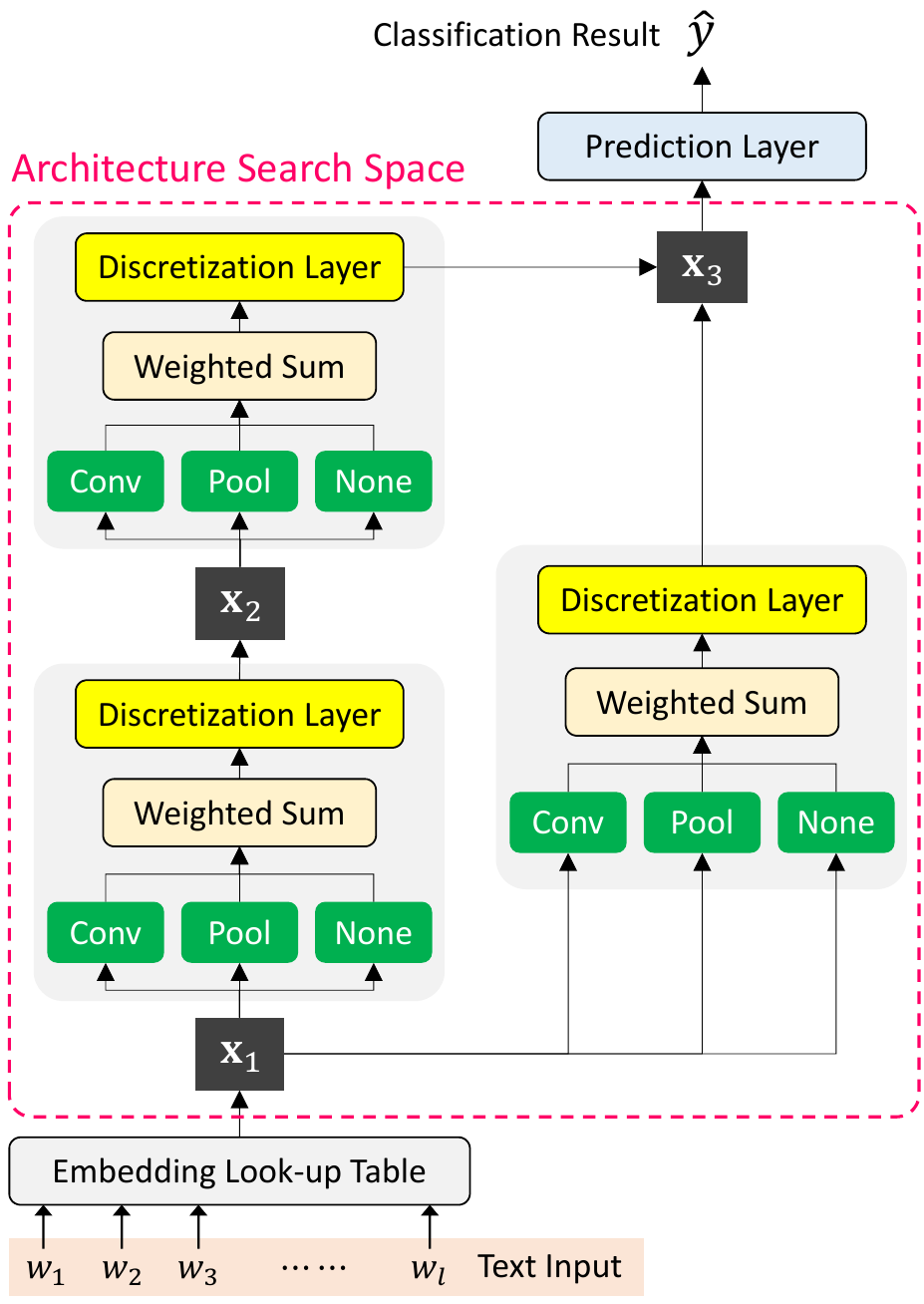}
\caption{The overview of the proposed DDNAS algorithm with a toy search space. A directed acyclic graph with three nodes is considered. Note that $\mathbf{x}_i$ is the latent representation produced by each layer. Green cells indicate three kinds of candidate operations, including convolution (Conv), pooling (Pool), and removing any operations (None).}
\label{fig-ovw}
\end{figure}

\section{The Proposed DDNAS Algorithm}
\label{sec-method}
We present the overview of the proposed DDNAS in Figure~\ref{fig-ovw}. Given a text document consisting of a sequence of words, we learn an embedding look-up table and retrieve the embedding of each word. The toy example of architecture search space contains three nodes of latent representations $\mathbf{x}_i$, and the candidate operations are put on edges. We aim at jointly learning the weighted combinations of operations and the parameters within each operation during the architecture search. Eventually, a prediction layer is learned to generate the classification result $\hat{y}$. The most important part of DDNAS is the discretization layer that takes action after deriving the embedding by the weighted sum of candidate operations. The discretization layer aims to map the continuous embedding into a latent discrete space so that the semantics of the input text can be better encoded. Discretized representations in the search space can also capture the potential hierarchical categorization of text data.

\subsection{Search Space}
\label{subsec-darts}
The input of DDNAS is a document $T$, e.g., news article or tweet, containing a sequence of words $T=(w_{1}, w_{2},...,w_{l})$, where $l$ is the number of words in the document. We feed the word sequence into a learnable embedding lookup table, and obtain the input representation vector, denoted as $\mathbf{x}_0\in\mathbb{R}^{d}$ with dimensionality $d$.

The proposed DDNAS model is based on the differentiable architecture search (DARTS)~\cite{darts19} to find the most proper model structure. We aim at learning a proper computation cell to be the building block of the resultant neural network model structure. A directed acyclic graph (DAG) $G=(V,E)$ with an ordered sequence of nodes is created to contain the search space. In the DAG, former nodes can extract lower-level features while latter ones are to capture higher-level semantics. Each node $v_i\in V$ is associated with a derived representation vector $\mathbf{x}_i$, and each directed edge $e_{i<j}\in E$ is associated with a certain transformation operation $o_{ij}\in \mathcal{O}$ (e.g., convolution) that maps $\mathbf{x}_i$ to $\mathbf{x}_j$, where $\mathcal{O}$ is the set of all candidate operations and $i<j$. Each latent representation $\mathbf{x}_j$ is the operation-wise sum over all of its predecessors, given by:
\begin{equation}
\label{eq-wsum}
\mathbf{x}_j=\sum_{i<j}o_{ij}(\mathbf{x}_i).
\end{equation}

By applying the softmax function to all candidate operations on each directed edge $e_{i<j}$ with learnable parameters, we can have a continuous search space, instead of selecting only one specific operation to transform $\mathbf{x}_i$. This can be depicted as:
\begin{equation}
\label{eq-darts}
\bar{o}_{ij}(\mathbf{x}_i) = \sum_{o\in \mathcal{O}}\frac{\exp(\alpha_{o}^{ij})}{\sum_{o'\in \mathcal{O}}\exp(\alpha_{o'}^{ij})}o(\mathbf{x}_i),
\end{equation}
where $\alpha_{o}\in\mathbb{R}^{|\mathbf{O}|}$ is the trainable weight vector, which can be updated via back propagation. We can consider that the transformed embedding of $\mathbf{x}_i$ is the sum of learnable weighted projections over all operations on $\mathbf{x}_i$. The learnable operation weight $\alpha_{o}^{ij}$ makes the neural architecture search differentiable. By optimizing the validation loss with gradient descent for all operation parameters $\alpha_{o}^{ij}$ and the weights within each operation, we can eventually obtain the final architecture by choosing the operation with the highest weight:
\begin{equation}
o_{ij}=\arg\max_{o\in\mathcal{O}}\alpha_{o}^{ij}.
\end{equation}

With the abovementioned search space, we need to determine the set of operations. We incorporate three categories of operation layers, including \textit{convolution}, \textit{pooling}, and \textit{none}. We choose these three due to the consideration of parallelization and the number of parameters. We do not consider \textit{recurrent} layers, such as the vanilla RNN~\cite{rnn95}, LSTM~\cite{lstm97} and GRU~\cite{gru15}, because they are hard to be parallelized and require more trainable parameters. Although the \textit{linear} layer allows parallel computing, it cannot well extract features from text data. Transformer~\cite{trm17} has a strong capability of textual feature extraction, and can support nice parallelization. However, it requires too many learnable parameters. For NAS, the operations should take time efficiency and space complexity into account. Hence, DDNAS utilizes only the following three categories of operations, and there are nine operations in total, whose detailed settings are summarized in Table~\ref{tab:optlist}. 
\begin{itemize}
\item \textit{Convolution.} We consider 1-D convolutional neural network to extract local features of texts like n-grams. Different filer sizes of CNN can also capture features with different granularities. Three kinds of 1-D CNN layers with filter sizes $3$, $5$, and $7$ are used. We use the convolution of $stride=1$ with padding to make the shape of output the same as the input. The number of output filters is the same as the input dimension. In addition, to capture long-range dependency between words, we also incorporate the dilated convolutional layer~\cite{dilate16} that introduces ``holes'' into each convolution filter. We consider three dilated steps, i.e., $2$, $4$, and $6$, with a fixed filter size $3$.
\item \textit{Pooling.} We utilize pooling since it can help train a deeper neural network~\cite{pool10}. Max-pooling and average-pooling are exploited. To keep the dimensionality the same before and after this layer, we again construct the pooling operations with padding and $stride=1$. We simply fix the filter size to $3$.
\item \textit{None.} This special operation is created to neglect some of the directed edges between nodes. We multiply zero to the input vector so that no information can be forwarded. None is an operation to avoid redundant or noisy information on edges. 
\end{itemize}

\begin{table}[!t]
\centering
\caption{List of adopted operations with their settings.}
\label{tab:optlist}
\resizebox{0.65\linewidth}{!}{%
\begin{tabular}{c|c|c|c}
\hline
 & Filter Size & Padding & Dilated Step \\ \hline\hline
Conv1D-3gram & 3 & 1 & \xmark \\ \hline
Conv1D-5gram & 5 & 2 & \xmark \\ \hline
Conv1D-7gram & 7 & 3 & \xmark \\ \hline
Conv1D-Dilated2 & 3 & 2 & 2 \\ \hline
Conv1D-Dilated4 & 3 & 4 & 4 \\ \hline
Conv1D-Dilated6 & 3 & 6 & 6 \\ \hline
Avg-Pooling & 3 & 1 & \xmark \\ \hline
Max-Pooling & 3 & 1 & \xmark \\ \hline
None & \xmark & \xmark & \xmark \\ \hline
\end{tabular}%
}
\end{table}

\subsection{Discretization Layer}
\label{subsec-dis}
Since text data is discrete in nature, we think it would be proper to discretize each latent representation $\mathbf{x}_i$. Discrete representations also have better interpretability than continuous embedding vectors. We can also consider the derived discrete variables as the latent topics of input text, which can benefit classification tasks. Inspired by Neural Bayes~\cite{nbayes20}, which allows learning representations by categorizing them via a generic parameterization, we aim at mapping each hidden representation $\mathbf{x}_i$ to a latent discrete space $z$ where the data distribution $p(\mathbf{x})$ is divided into a fixed number of arbitrary conditional distributions. We can discretize each derived hidden representation $\mathbf{x}_i$ by maximizing mutual information between observed random variables (i.e., $\mathbf{x}$) and latent discrete random variable $z$.

Let the latent discrete space be $z$, and the output from the discretization layer is $\mathbf{x}$. Assume that the data distribution $p(\mathbf{x})$ is conditioned on a discrete random variable $z$ with $K$ states. The probability of $z$ belonging to the $k$-th state, denoted as $p(z=k)$, can be estimated through the expectation of $L_k(\mathbf{x})$, which is a hidden layer that we perform the softmax function and obtain the discretization probability on state $k$. We can treat function $L$ as a soft categorization function accepting input representation $\mathbf{x}$.

We can leverage Bayes' rule to depict the parameterization of the discretization layer. Let $p(\mathbf{x}|z=k)$ and $p(z)$ be any conditional and marginal distributions, respectively, defined for continuous random variable $\mathbf{x}$ and discrete random variable $z$. If the expectation $\mathbb{E}_{\mathbf{x}\sim p(\mathbf{x})}\left[L_k(\mathbf{x}) \right] \neq 0$,
$\forall k \in [K]$, then we can derive a mapping function $L(\mathbf{x}): \mathbb{R}^n\rightarrow \mathbb{R}^{+^K}$ 
for any given input data $\mathbf{x} \in \mathbb{R}^n$, where $n$ is the number of data samples and $\sum_{k=1}^K L_k(\mathbf{x})=1$, $\forall \mathbf{x}$, such that:
\begin{equation}
\label{eq-lemma}
\begin{aligned}
p(\mathbf{x}|z=k) &= \frac{p(\mathbf{x},z=k)}{\int p(\mathbf{x},z=k) \mathrm{d}\mathbf{x}}\\
&= \frac{p(z=k|\mathbf{x})p(\mathbf{x})}{\int p(z=k|\mathbf{x})p(\mathbf{x}) \mathrm{d}\mathbf{x}}\\
&= \frac{L_k(\mathbf{x})\cdot p(\mathbf{x})}{\mathbb{E}_{\mathbf{x}\sim p(\mathbf{x})}\left[L_k(\mathbf{x}) \right]}
\end{aligned}
\end{equation}
where $p(z=k)=\mathbb{E}_{\mathbf{x}}\left[L_k(\mathbf{x}) \right]$ and $p(z=k|\mathbf{x})=L_k(\mathbf{x})$. Since the expectation $L$ is calculated via a neural network layer, we can replace $L$ with $L_{\theta}$, in which $\theta$ denotes the learnable parameters.

We aim to generate the latent discrete space $z$ with $K$ states ($K=32$ by default) for the data distribution $p(\mathbf{x})$ such that the mutual information between $\mathbf{x}$ and $z$ is maximized. Maximizing mutual information~\cite{mim20} would enforce $z$ to utilize only $K$ discrete states to cover the knowledge in the continuous representation $\mathbf{x}$ as maximum as possible. Specifically, based on the Neural Bayes parameterization~\cite{nbayes20}, we aim to find the optimal $L*$ to maximize mutual information between $\mathbf{x}$ and $z$, given by:
\begin{equation}
\label{eq-mim}
L^{*} = \arg\max_{L}\mathbb{E}_{\mathbf{x}}\left[\sum_{k=1}^{K}L_{k}(\mathbf{x})\log\frac{L_{k}(\mathbf{x})}{\mathbb{E}_{\mathbf{x}}\left[L_{k}(\mathbf{x})\right]}\right].
\end{equation}
Based on $L^{*}$, we can derive the conditional distribution $p(z|\mathbf{x})$, and accordingly unfold which discrete state of $z$ can be the semantic representative of $\mathbf{x}$ for text classification.

Now we can formulate the final objective of learning discrete representation for $\mathbf{x}$. The equation Eq.~\ref{eq-mim} of finding $L^{*}$ is parameterized by $\theta$. Hence, we can take negative into Eq.~\ref{eq-mim} and rewrite it as a minimization task, i.e., finding $\theta$ that minimizes: 
\begin{equation}
-\mathbb{E}_\mathbf{x}[\sum_{k=1}^K L_{\theta_k}(\mathbf{x})\log L_{\theta_k}(\mathbf{x})]+\sum_{k=1}^K\mathbb{E}_{\mathbf{x}}[L_k(\mathbf{x})]\log\mathbb{E}_{\mathbf{x}}[L_k(\mathbf{x})].
\end{equation}
To have a better capability of discrete distributed representation learning and to improve the model interpretability, we can use cross-entropy to replace the second term (i.e., negative-entropy), resulting in the final learning objective of discretization $J_D(\theta)$:
\begin{equation}
\label{eq-objdis}
\begin{aligned}
J_{D}(\theta)=&-\mathbb{E}_{\mathbf{x}}\left[\sum_{k=1}^{K}L_{\theta_{k}}(\mathbf{x})\log\left(L_{\theta_{k}}(\mathbf{x}\right) \right] \\
&-\sum_{k=1}^{K}\frac{1}{K}\log \left(\mathbb{E}_{\mathbf{x}}\left[L_{\theta_{k}}(\mathbf{x})\right]\right)\\ 
&+ \frac{K-1}{K}\log\left(1-\mathbb{E}_{\mathbf{x}}[L_{\theta_{k}}(\mathbf{x})]\right).
\end{aligned}
\end{equation}
This discretization objective is imposed into every derived latent representation $\mathbf{x}_j$, mentioned in Equation Eq.~\ref{eq-wsum}, i.e., after each directed edge of operation in the model architecture. In other words, we will be allowed to learn and exploit a sort of hierarchical latent feature categorization of input text, from lower- to higher-level discretization in the DAG, to boost the performance of text classification. We will show how the proposed discretization layer captures the hierarchical latent categorization via visualization in the experiment. Note that the discretization layer with $J_D(\theta)$ can be also seen as a novel kind of regularization in NAS.

\subsection{Joint Learning}
\label{subsec-opt}
We aim at generating the classification results by utilizing the derived embeddings of all words in the given text. After obtaining the latent representation $\mathbf{x}_w$ of word $w$ from the last node in the DAG, we use a linear layer to produce its embedding $\dot{\mathbf{x}}_w$, given by: $\dot{\mathbf{x}}_w=\mathbf{x}_w\mathbf{W}_1+\mathbf{b}_1$, where $\mathbf{W}_1$ and $\mathbf{b}_1$ are learnable parameters. Then we sum over the embeddings of all words to generate the embedding $\mathbf{x}_T$ of the given text $T$: $\mathbf{x}_T=\sum_{w\in T} \dot{\mathbf{x}}_w$. Last, by feeding $\mathbf{x}_T$ into a hidden layer, we utilize a softmax function to generate the prediction outcome of being class $c$, given by:
\begin{equation}
\hat{y}_c = \sigma(\mathbf{x}_T\mathbf{W}_T+\mathbf{b}_T),
\end{equation}
where $\mathbf{W}_T$ and $\mathbf{b}_T$ are trainable weights, and $\sigma$ is the softmax function. We use Cross-Entropy to calculate the loss between predicted class probability $\hat{y}_c$ and the ground-truth class $y_c$, given by:
\begin{equation}
\label{eq-cls}
J_C(\theta) = -\frac{1}{N}\sum_{c=1}^C\left[ y_c\log \hat{y}_c + (1-y_c)\log(1-\hat{y}_c) \right],
\end{equation}
where $C$ is the number of classes and $N$ is the number of data instances.

Our ultimate learning target of NAS is to not only maximize the classification accuracy by minimizing $J_C(\theta)$, but also to maximize the mutual information between data distribution and discretized representation by minimizing $J_D(\theta)$. We simultaneously optimize such two objectives, and the final objective is given by:
\begin{equation}
\label{eq-finalobj}
J(\theta) = \lambda\cdot J_D(\theta) + (1-\lambda)\cdot J_C(\theta),
\end{equation}
where $\lambda$ is a balancing hyperparameter that determines the importance of $J_D$ and $J_C$. We utilize the Adam optimizer~\cite{adam15} to learn $\theta$ as it can determine the learning rate abortively.

\textbf{Learning Latent Hierarchical Categorization.} In the proposed DDNAS, the search of neural architecture is over the space depicted by a directed acyclic graph (DAG). Hence, the obtained neural architecture can be considered as a kind of latent hierarchical categorization for the input text data. Here we aim at explaining the idea of hierarchical categorization and justify the discretization layer that leads to such categorization. The directed acyclic graph had been utilized to depict the hierarchical categorization in the literatures~\cite{hiersuv11,costa2007review,feng2022hierarchical,HUANG2020105655}. We realize the hierarchical categorization through two designs in the proposed DDNAS. First, DDNAS leverages and creates DAG to depict the neural search space. In the DAG, the early search nodes can represent finer-grained categorization while the latter search nodes can capture coarser-grained categorization. Second, since the categorization itself is in a discrete structure, we propose and impose the discretization layer within each search node in DDNAS. The distribution of discretized states within a search node can be treated as a certain categorization. By combining DAG with the discretization layer on each search node, the search for a proper neural architecture for text data can be seen as the learning of latent hierarchical categorization.

\section{Experiments}
\label{sec-exp}
We aim to answer the following evaluation questions. 
\begin{itemize}
\item \textbf{EQ1}: Can our DDNAS model achieve satisfactory performance on text classification tasks when compared to existing state-of-the-art NAS-based methods? 
\item \textbf{EQ2}: How does each type of operation in DDNAS contribute to the performance? 
\item \textbf{EQ3}: Can DDNAS produce accurate classification outcomes on diverse types of texts? 
\item \textbf{EQ4}: How do different hypermeters of DDNAS affect the classification performance? 
\item \textbf{EQ5}: How does the searched architecture look like?
\item \textbf{EQ6}: Can different classes be captured by latent hierarchical categorization via discretization?
\item \textbf{EQ7}: Can DDNAS learn effective model architectures with a smaller training set?
\end{itemize}

\begin{table}[!t]
\centering
\caption{List of used datasets with their statistics.}
\label{tab-data}
\resizebox{.6\linewidth}{!}{%
\begin{tabular}{c|c|c|c}
\hline
Data & Category & \#Instances & \#Class \\ \hline\hline
IMDB & Movie Reviews & 50,000 & 2 \\ \hline
AG & News Articles & 127,600 & 4 \\ \hline
Yelp & Business Reviews & 598,000 & 2 \\ \hline
Instagram & Tweets & 2,218 & 2 \\ \hline
Vine & Tweets & 970 & 2 \\ \hline
Twitter & Tweets & 483 & 2 \\ \hline
NYT & News Articles & 13,081 & 5 \\ \hline
20News & News Articles & 18,846 & 6 \\ \hline
\end{tabular}%
}
\end{table}

\subsection{Datasets \& Evaluation Settings}
\label{subsec-expset}
\textbf{Datasets.} 
We utilize eight well-known datasets in the experiments. The data statistics are provided in Table~\ref{tab-data}. These datasets can be divided into three diverse categories: three belong to news articles, two are about item reviews, and three are related to user-generated short texts (i.e., tweets).
\begin{enumerate}
\item \textit{IMDB Movie Review} data\footnote{\url{https://www.kaggle.com/lakshmi25npathi/imdb-dataset-of-50k-movie-reviews}}~\cite{wvsent}: classifying sentiments with binary class, i.e., ``positive'' and ``negative'', based on IMDB text reviews of movies.
\item \textit{AG News Article} data\footnote{\url{https://www.kaggle.com/amananandrai/ag-news-classification-dataset}}~\cite{charcnn}: a text classification benchmark for classifying news articles with four classes (news topics), i.e., ``world'', ``sports'', ``business'', and ``science.''
\item \textit{Yelp Business Review} data\footnote{\url{https://www.yelp.com/dataset/}}: predicting the review polarity with ``positive'' and ``negative'' classes, and the dataset is obtained from the Yelp Dataset Challenge in 2015.
\item \textit{Instagram} data\footnote{\label{note1}\url{https://sites.google.com/site/cucybersafety/home/cyberbullying-detection-project/dataset}}~\cite{igdata15}: perform cyberbullying detection on Instagram tweets with binary classification, i.e., ``bullying'' or ``normal.''
\item \textit{Vine} data\footref{note1}~\cite{vinedata15}: predicting whether a tweet on a mobile application platform Vine is bullied by any commenter with binary class, i.e., ``bullying'' or ``normal.''
\item \textit{Twitter} data\footnote{\url{https://www.dropbox.com/s/7ewzdrbelpmrnxu/rumdetect2017.zip?dl=0}}~\cite{twdata16}: detecting fake messages on tweets with binary classification, i.e., ``true'' or ``fake.''
\item \textit{NYT} data\footnote{https://github.com/dheeraj7596/ConWea/tree/master/data}: classifying news articles published by The New York Times into five genres, such as ``arts'' and ``sports.''
\item \textit{20News}: data\footnote{\url{http://qwone.com/~jason/20Newsgroups/}}: classifying newsgroup documents into six categories, such as ``computer'', ``recreation'', ``science.''
\end{enumerate}


\textbf{Competing Methods.}
We compare our DDNAS with the state-of-the-art method and several baselines, as listed below.
\begin{itemize}
\item \textbf{C-LSTM}~\cite{clstm15}: stacking convolutional layers with long-short term memory (LSTM) for text classification.
\item \textbf{Transformer}~\cite{trm17}: multi-head self-attention layers followed by one or more feed-forward layers, and $6$-layers transformer is adopted. 
\item \textbf{ENAS}~\cite{enas18}: an efficient NAS based on a training scheme with shared parameters among architectures, instead of fully training each structure individually.
\item \textbf{DARTS}~\cite{darts19}: a differentiable architecture search model that jointly optimizes model weights and architecture weights using gradient descent, and it is the basis of our DDNAS.
\item \textbf{TextNAS}~\cite{textnas20}: the state-of-the-art NAS-based model tailored for text representation learning and classification.
\end{itemize}
Unless specified, we use the default settings in their open-source codes without tuning their hyperparameters or modifying the proposed search spaces except for replacing all 2-D convolutions with 1-D.

Note that pre-trained transformer-based models, such as BERT~\cite{bert} and RoBERTa~\cite{roberta}, are popular and powerful for text classification. And one should compare a new proposed pre-trained method with pre-trained transformer-based models. However, we think it is unfair to experimentally compare with pre-trained transformer-based models in evaluating the effectiveness of a neural architecture search (NAS) based method like our proposed DDNAS. The reason is two-fold. First, the NAS-based methods train from scratch to find the effective model architecture, i.e., without relying on external datasets for the search. Although fine-tuning pre-trained models can usually be performed in a lightweight manner (i.e., on small-scale data), they require a large amount of external data to have effective model pre-training. Because the learning of NAS-based methods and pre-trained methods is based on different data settings, the selection of our baselines follows NAS-based studies~\cite{sdarts20,pdarts19,sgas20,fenas20,autoatt21}, which compare with only NAS methods and do not compare to pre-trained models. Second, our evaluation goal is to examine whether the proposed DDNAS can improve existing NAS-based methods for text classification. Hence, we mainly focus on employing state-of-the-art text NAS as the competing methods. Since the approaches between NAS and pre-trained models are essentially different, and comparing them to pre-trained models cannot inform us where the performance improvement comes from in the line of text NAS research, we do not compare DDNAS with pre-trained transformer-based methods.

\textbf{Model Configuration.}
For our model and all competing methods, by default, we set the batch size as $64$, the dimension of hidden representation as $256$, the maximum input length as $384$ with zero padding (i.e., every document is padded and appended with zeros so that the lengths of all documents are with the same input size $384$), and the number of nodes in the search graph $G$ is $5$ (for DDNAS and NAS-based competitors, including ENAS, DARTS, and TextNAS). In addition, the number of states $K$ for discrete random variables $z$ is set as $64$. The initial learning rate in Adam optimizer is set as $0.001$. The hyperparameters of competing methods are set by following the settings mentioned in respective studies. The default setting of $\lambda$ in DDNAS is $0.5$. All experiments are conducted with PyTorch running on GPU machines (NVIDIA Tesla V100 32GB).

\textbf{Metrics \& Settings.}
The evaluation metrics include Accuracy (Acc) and F1 scores. We follow the split percentage of training and testing in the original datasets for our experiments. For those datasets with the original split setting, we compile two evaluation sets with different splittings. One is randomly sampling 80\% for training and the remaining 20\% for testing, which is the default setting, while the other is randomly sampling 20\% for training and the remaining 80\% for testing. The latter splitting is utilized to examine whether a NAS-based text classification model can still work effectively with fewer training samples. A good NAS-based text classification model should not totally rely on the complicated architecture searched from a massive training dataset. One should expect an effective but compact model architecture learned and generated from a limited number of training samples. For those datasets without specifying the validation set, we randomly select 5\% samples from the training set as validation data. To determine the model architecture for the experiments, we follow DARTS~\cite{darts19} to run DDNAS ten times with different random initialization seeds, and pick the best one based on the validation performance. To evaluate the derived model architecture, we do not use the weights learned during the search process; instead, we randomly initialize all of its learnable parameters. That said, we train the model based on the final architecture from scratch, and report its performance on the test set. The test set is confirmed to never be utilized for architecture search or architecture selection. The conducted train-test is repeated $20$ times, and the average performance scores are reported.

\begin{table*}[!t]
\centering
\caption{Main experimental results across eight datasets and six compared methods under the setting of 80\% training and 20\% testing data. The best model and the best competitor are highlighted by \textbf{bold} and \underline{underline}, respectively. For each dataset, we conduct significance test against the best reproducible model, and the symbol \textbf{*} means that the improvement is significant at a 0.05 significance level.}
\label{exp-mainres}
\resizebox{1.0\textwidth}{!}{%
\begin{tabular}{c|c|c|c|c|c|c|c|c}
\hline
 & \multicolumn{2}{c|}{IMDB} & \multicolumn{2}{c|}{AG News} & \multicolumn{2}{c|}{Yelp} & \multicolumn{2}{c}{Instagram} \\ \hline
 & Acc & F1 & Acc & F1 & Acc & F1 & Acc & F1  \\ \hline
C-LSTM & 0.839 & \underline{0.837} & 0.805 & 0.797 & 0.869 & 0.865 & 0.704 & 0.527  \\ \hline
Transformer & 0.825 & 0.805 & 0.838 & 0.826 & 0.852 & 0.854 & 0.783 & \underline{0.677}  \\ \hline
ENAS & 0.821 & 0.779 & \underline{0.851} & 0.827 & 0.859 & 0.844 & \underline{0.820} & 0.630  \\ \hline
DARTS & 0.843 & 0.827 & 0.843 & \underline{0.851} & \underline{0.875} & \underline{0.888} & 0.750 & 0.636 \\ \hline
TextNAS & \underline{0.863} & 0.825 & 0.812 & 0.793 & 0.863 & 0.825 & 0.806 & 0.658  \\ \hline
DDNAS & \textbf{0.887*} & \textbf{0.872*} & \textbf{0.895*} & \textbf{0.892*} & \textbf{0.886*} & \textbf{0.896*} & \textbf{0.822*} & \textbf{0.692*} \\ \hline\hline

 & \multicolumn{2}{c|}{Vine} & \multicolumn{2}{c|}{Twitter} & \multicolumn{2}{c|}{NYT} & \multicolumn{2}{c}{20News} \\ \hline
 & Acc & F1 & Acc & F1 & Acc & F1 & Acc & F1 \\ \hline
C-LSTM & 0.639 & 0.435 & 0.804 & 0.808 & 0.851 & 0.782 & 0.692 & 0.653 \\ \hline
Transformer & 0.706 & 0.495 & 0.958 & \underline{0.963} & 0.882 & 0.821 & 0.738 & 0.672 \\ \hline
ENAS & \underline{0.803} & 0.503 & 0.929 & 0.688 & 0.901 & 0.848 & 0.745 & 0.693 \\ \hline
DARTS & 0.718 & \underline{0.689} & 0.812 & 0.800 & 0.922 & 0.884 & 0.752 & 0.713 \\ \hline
TextNAS & 0.799 & 0.551 & \underline{0.976} & 0.727 & \underline{0.951} & \underline{0.902} & \underline{0.761} & \underline{0.754} \\ \hline
DDNAS & \textbf{0.819*} & \textbf{0.698*} & \textbf{0.979} & \textbf{0.982*} & \textbf{0.979*} & \textbf{0.932*} & \textbf{0.807*} & \textbf{0.767*} \\ \hline
\end{tabular}%
}
\end{table*}

\begin{table*}[!t]
\centering
\caption{Main experimental results across eight datasets and six compared methods under 20\% training and 80\% testing.}
\label{exp-ressmall}
\resizebox{1.0\textwidth}{!}{%
\begin{tabular}{c|c|c|c|c|c|c|c|c}
\hline
 & \multicolumn{2}{c|}{IMDB} & \multicolumn{2}{c|}{AG News} & \multicolumn{2}{c|}{Yelp} & \multicolumn{2}{c}{Instagram} \\ \hline
 & Acc & F1 & Acc & F1 & Acc & F1 & Acc & F1  \\ \hline
C-LSTM & 0.811 & 0.792 & 0.701 & 0.681 & 0.841 & 0.821 & 0.682 & 0.507  \\ \hline
Transformer & 0.791 & 0.781 & \underline{0.758} & 0.724 & 0.831 & 0.811 & 0.753 & 0.645  \\ \hline
ENAS & 0.802 & 0.789 & 0.761 & \underline{0.734} & 0.839 & 0.823 & \underline{0.798} & \underline{0.665}  \\ \hline
DARTS & 0.801 & 0.788 & 0.722 & 0.701 & \underline{0.848} & \underline{0.842} & 0.741 & 0.632 \\ \hline
TextNAS & \underline{0.831} & \underline{0.812} & 0.702 & 0.693 & 0.842 & 0.805 & 0.780 & 0.632  \\ \hline
DDNAS & \textbf{0.842*} & \textbf{0.826*} & \textbf{0.780*} & \textbf{0.779*} & \textbf{0.866*} & \textbf{0.860*} & \textbf{0.811*} & \textbf{0.667*} \\ \hline\hline

 & \multicolumn{2}{c|}{Vine} & \multicolumn{2}{c|}{Twitter} & \multicolumn{2}{c|}{NYT} & \multicolumn{2}{c}{20News} \\ \hline
 & Acc & F1 & Acc & F1 & Acc & F1 & Acc & F1 \\ \hline
C-LSTM & 0.419 & 0.302 & 0.502 & 0.583 & 0.703 & 0.691 & 0.562 & 0.551 \\ \hline
Transformer & 0.502 & 0.354 & 0.542 & 0.612 & 0.741 & 0.723 & \underline{0.621} & \underline{0.582} \\ \hline
ENAS & \underline{0.612} & 0.434 & 0.568 & 0.633 & 0.851 & \underline{0.813} & 0.572 & 0.561 \\ \hline
DARTS & 0.578 & 0.421 & 0.558 & 0.623 & 0.911 & 0.702 & 0.581 & 0.562 \\ \hline
TextNAS & 0.602 & \underline{0.475} & \underline{0.602} & \underline{0.712} & \underline{0.943} & 0.713 & 0.605 & 0.573 \\ \hline
DDNAS & \textbf{0.613*} & \textbf{0.576*} & \textbf{0.639*} & \textbf{0.720*} & \textbf{0.961*} & \textbf{0.825*} & \textbf{0.625*} & \textbf{0.612*} \\ \hline
\end{tabular}%
}
\end{table*}

\subsection{Experimental Results}
\label{subsec-expres}
\textbf{Main Results.} 
The main experimental results are shown in Table~\ref{exp-mainres}, which answers \textbf{EQ1} and \textbf{EQ3}. We can find that the proposed DDNAS consistently outperforms the state-of-the-art NAS-based method TextNAS and other baselines over two metrics across eight datasets. By looking into the details, we can obtain three findings. First, since the performance of DDNAS is significantly better than TextNAS, it shows the effectiveness of discretized representation learning and differentiable neural architecture search. Second, while the main difference between DDNAS and DARTS is the discretization layer, DDNAS exhibits outstanding performance improvement. Such a result directly proves the usefulness of discretization in text representation learning, and indirectly verifies the effectiveness of learning latent hierarchical categorization. Third, among all datasets, it can be observed that the superiority of DDNAS on short-text datasets, especially Instagram and Twitter tweets, is relatively small. A possible reason is that the scale of these datasets is small, and thus all NAS-based methods do not have enough data to learn a better neural architecture.

\textbf{Results with Fewer Training Data.}
To answer \textbf{EQ7}, i.e., understand whether DDNAS can learn with a limited number of training instances, we conduct the experiments with 20\% training and 80\% testing for each dataset. The results are exhibited in Table~\ref{exp-ressmall}. It can be again observed that DDNAS leads to the best performance across all datasets and metrics. Such results deliver an important message: DDNAS is able to not only produce effective model architectures, but also learn them without requiring too much training data. We think the reason comes from the fact that DDNAS does not rely on complicated operations, such as LSTM~\cite{lstm97} and Transformer~\cite{trm17}, as its building blocks. NAS with LSTM and Transformer contains too many learnable parameters, and thus requires more training data. Given that DDNAS can learn from limited training data and bring promising performance, it would be useful for real-world text classification applications whose labels are costly to collect.

\begin{figure}[!t]
\centering
\includegraphics*[width=1.0\linewidth]{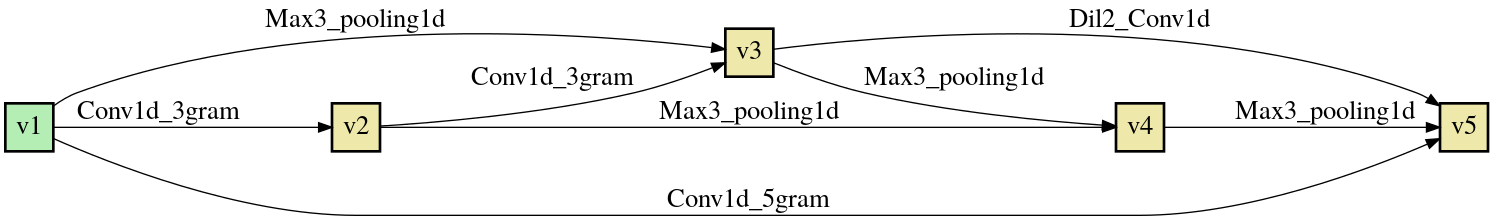}
\caption{Searched $5$-nodes architecture on IMDB.}
\label{fig-exarch}
\end{figure}
\begin{figure}[!t]
\centering
\includegraphics*[width=0.875\linewidth]{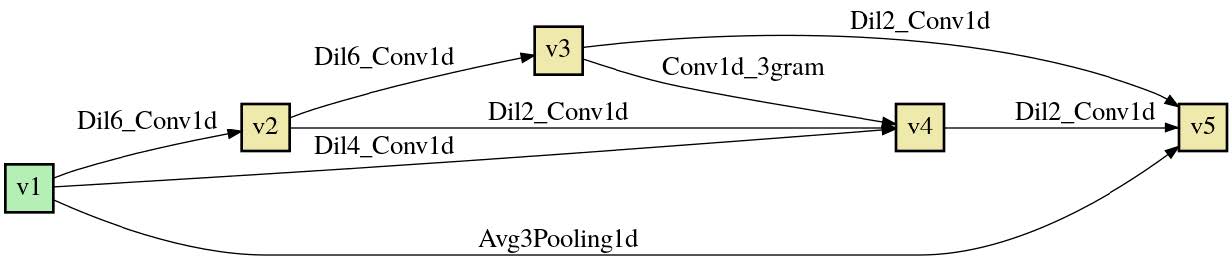}
\caption{Searched $5$-nodes architecture on AG News.}
\label{fig-exarch2}
\end{figure}

\textbf{Architecture Visualization.}
To answer \textbf{EQ5}, we demonstrate the model architectures searched by DDNAS using IMDB and AG News datasets in Figure~\ref{fig-exarch} and Figure~\ref{fig-exarch2}, respectively, in which the number of representation nodes in DAG search space is $5$. It is apparent that the discovered architectures are not common in neural network structures, and have a few interesting properties. First, there is a highway that directly transmits the lowest-level information (fine-grained features) to the final representation. This highway can be regarded as a kind of residual connection that benefits the training of deeper networks. Second, the discovered architectures contain some pooling layers to distill significant features and neglect noisy information. Third, there exist consecutive convolution layers, e.g., $v_1\rightarrow v_2\rightarrow v_3\rightarrow v_5$, to form deep networks so that the model can gradually extract low-level features to assemble high-level semantic knowledge. Last, ``None'' operations do take effect on removing directional connections, e.g., $v_2 \rightarrow v_5$.

\begin{figure}[!t]
\centering
\includegraphics*[width=1.0\linewidth]{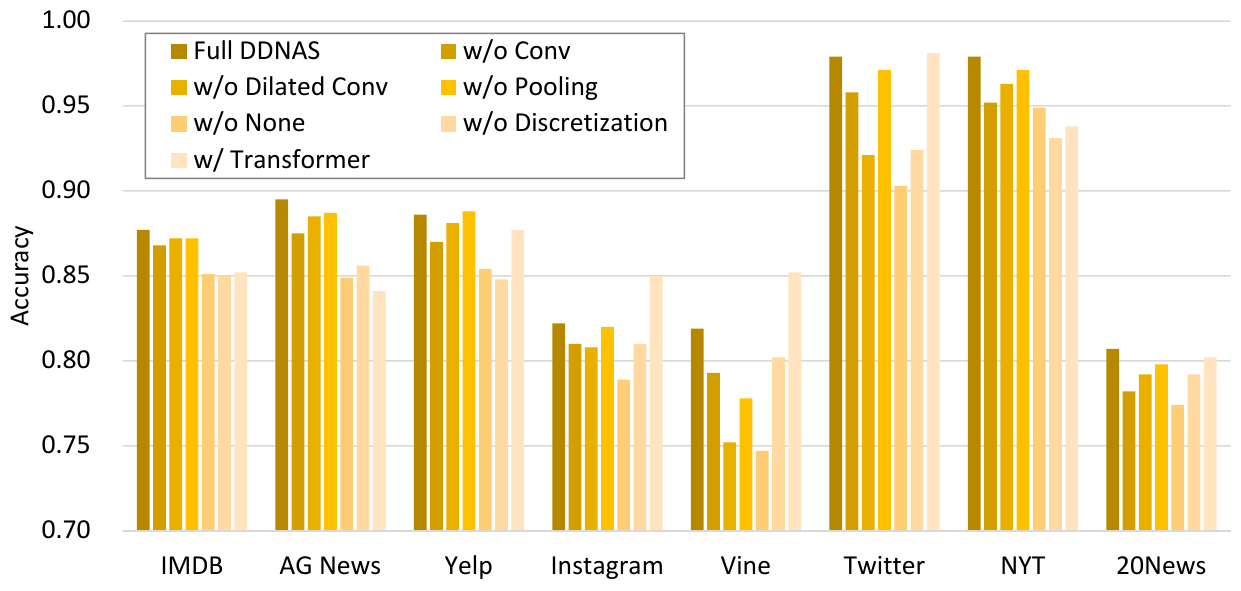}
\caption{Ablation Study: accuracy of the original DDNAS (Full DDNAS) removing each operation type, the discretization layer (w/o Discretization), and adding the Transformer operation (w/ Transformer).}
\label{fig-exabla}
\end{figure}

\begin{figure}[!t]
\centering
\includegraphics*[width=0.75\linewidth]{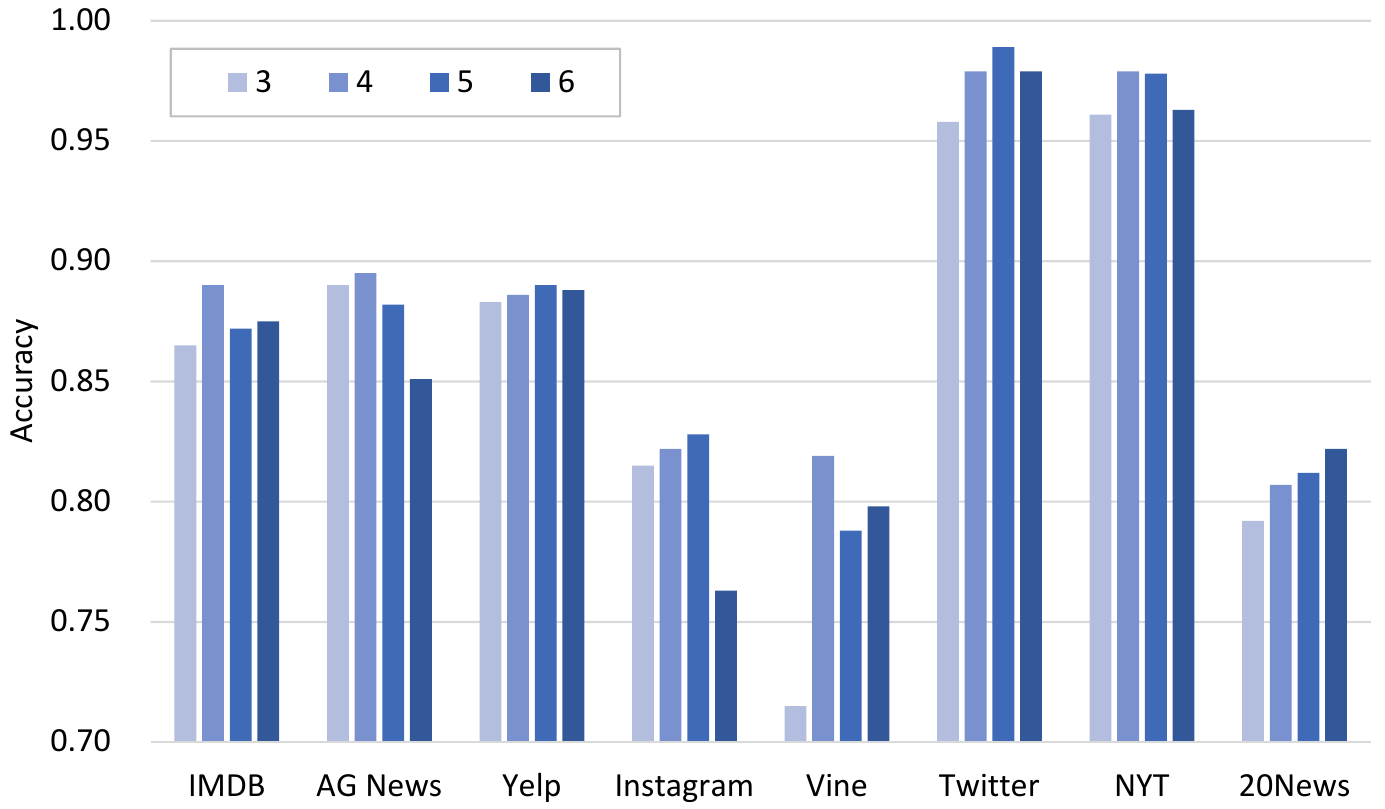}
\caption{Accuracy by varying \# of nodes in DAG.}
\label{fig-expnode}
\end{figure}

\subsection{Model Analysis}
\label{subsec-expana}
\textbf{Ablation Study.}
To answer \textbf{EQ2}, we report how each category of DDNAS operations contributes to the performance by removing each one from the entire method. What we aim to study includes Convolution (Conv), Dilated Conv., Pooling, and None. The results in Accuracy on eight datasets are presented in Figure~\ref{fig-exabla}. We can find every operation indeed plays a certain contribution. It is especially important on None and Conv because removing either brings a significant performance drop. Such results exhibit the necessity of feature extraction by 1-D convolution mechanisms, and properly discarding redundant and noisy information passing can better learn text representations. The pooling operation has limited contribution to the architecture. Since pooling does not involve learnable parameters, it could have a weaker capability in producing precise representations of hierarchical categorization. 

To understand where the performance improvement comes from, we conduct additional experiments in the ablation study. Since we wonder whether the improvement comes from the use of the discretization layer or from removing the Transformer operation, we add two model variants of DDNAS in the ablation study. One is removing the discretization layer (w/o Discretization), and the other is adding the Transformer operation (w/ Transformer). The results are exhibited in Figure~\ref{fig-exabla}. We can find that the DDNAS variant without the discretization layer consistently hurts the performance. The reduced accuracy also verifies the usefulness of the latent hierarchical categorization learned with the discretization layer. Besides, the DDNAS variant with the Transformer operation as one of the building blocks tends to lower the accuracy scores, i.e., in six out of eight datasets. DDNAS with Transformer outperforms the original DDNAS in only two datasets (i.e., Instagram and Vine). Such results bring two insights. First, DDNAS with basic operations (without Transformer) is enough to produce promising performance in most datasets. A strong operation like Transformer could weaken the role of basic operations, and thus damage the performance. Second, for datasets with short texts, like Instagram, Vine, and Twitter, DDNAS with Transformer has the potential to improve performance. This may result from the fact that the effective representation learning of short texts in social media requires deeper and longer correlation modeling between words~\cite{surprise21www,dltextcls1,dltextcls2}, which is one of the strengths of Transformer-based models. 

\begin{figure}[!t]
\centering
\includegraphics*[width=0.7\linewidth]{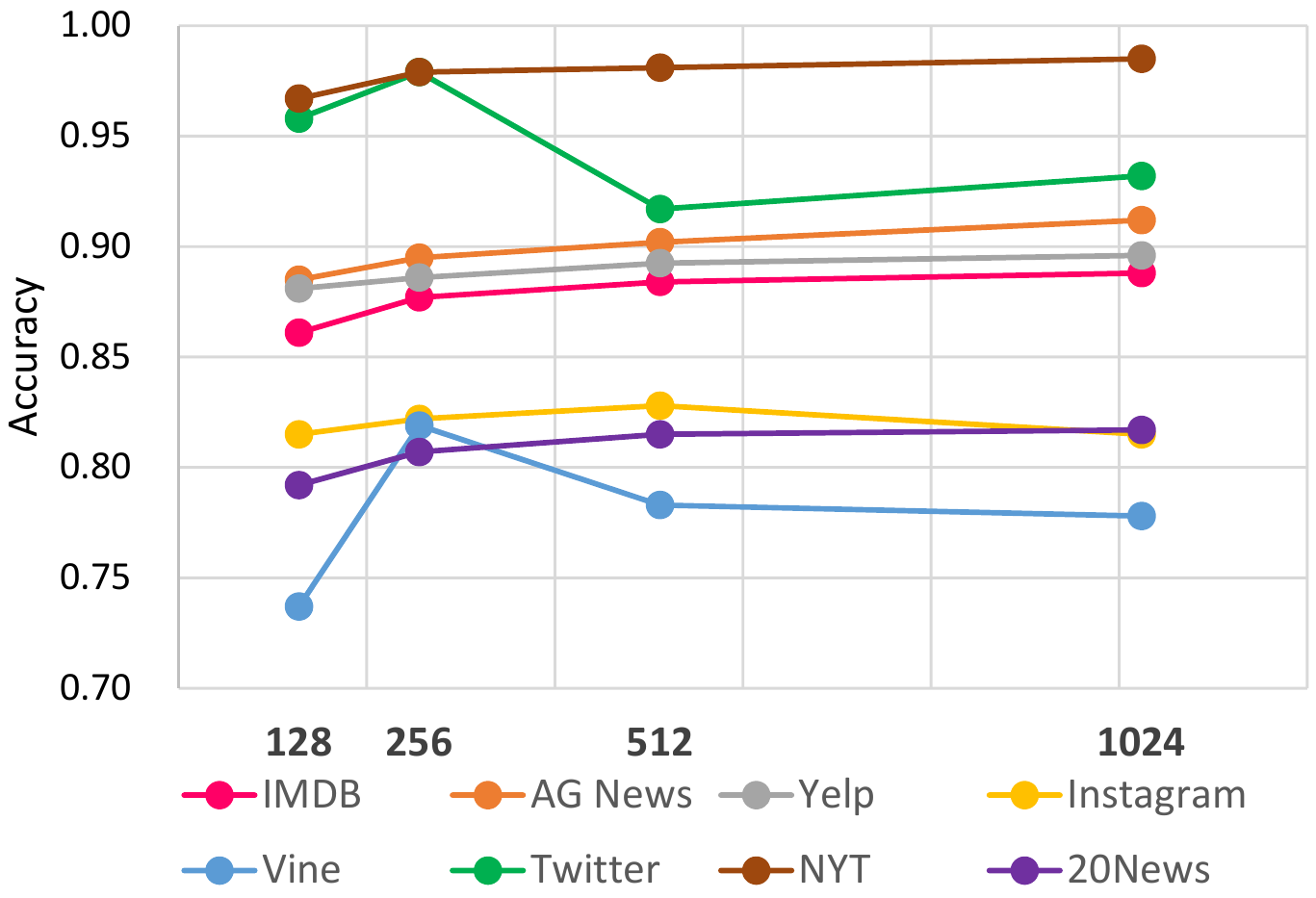}
\caption{Accuracy by varying embedding dimension.}
\label{fig-expdim}
\end{figure}

\textbf{Number of Representation Nodes.}
To understand how the complexity of model architecture affects the classification performance, we vary the number of representation nodes in the DAG search space. By varying the number of nodes as $3$, $4$, $5$, and $6$, we present the results on eight datasets in Figure~\ref{fig-expnode} that also answers \textbf{EQ4}. We can find the architectures with $4$ and $5$ nodes lead to higher accuracy scores. Too simple architectures e.g., $3$ nodes, are hard to encode the semantics of input text. Too complicated model architectures, e.g., $6$ nodes, could bring the overfitting issue in NAS methods. Hence, for text classification, we would suggest having $4$ or $5$ nodes to design the search space of DDNAS.

\textbf{Representation Dimensionality.}
We examine how the dimensionality of text representations affects performance. We show the results by setting the dimensionality as $128$, $256$, $512$, and $1024$ in Figure~\ref{fig-expdim}, which also answers \textbf{EQ4}. Two findings can be derived from the results. First, for datasets with sufficient numbers of instances (i.e., IMDB, AG, Yelp, and Instagram), the performance gets gradually improved when the dimensionality is increased. The best results locate at $1024$. 
Second, for datasets with limited numbers of instances (i.e., Vine and Twitter), the accuracy scores are high at $256$, but become low when too large dimensionality is utilized. Such a result is reasonable because larger dimensionalities require more training instances to generate better representations. We suggest setting $256$.

\begin{figure}[!t]
\centering
\includegraphics*[width=0.7\linewidth]{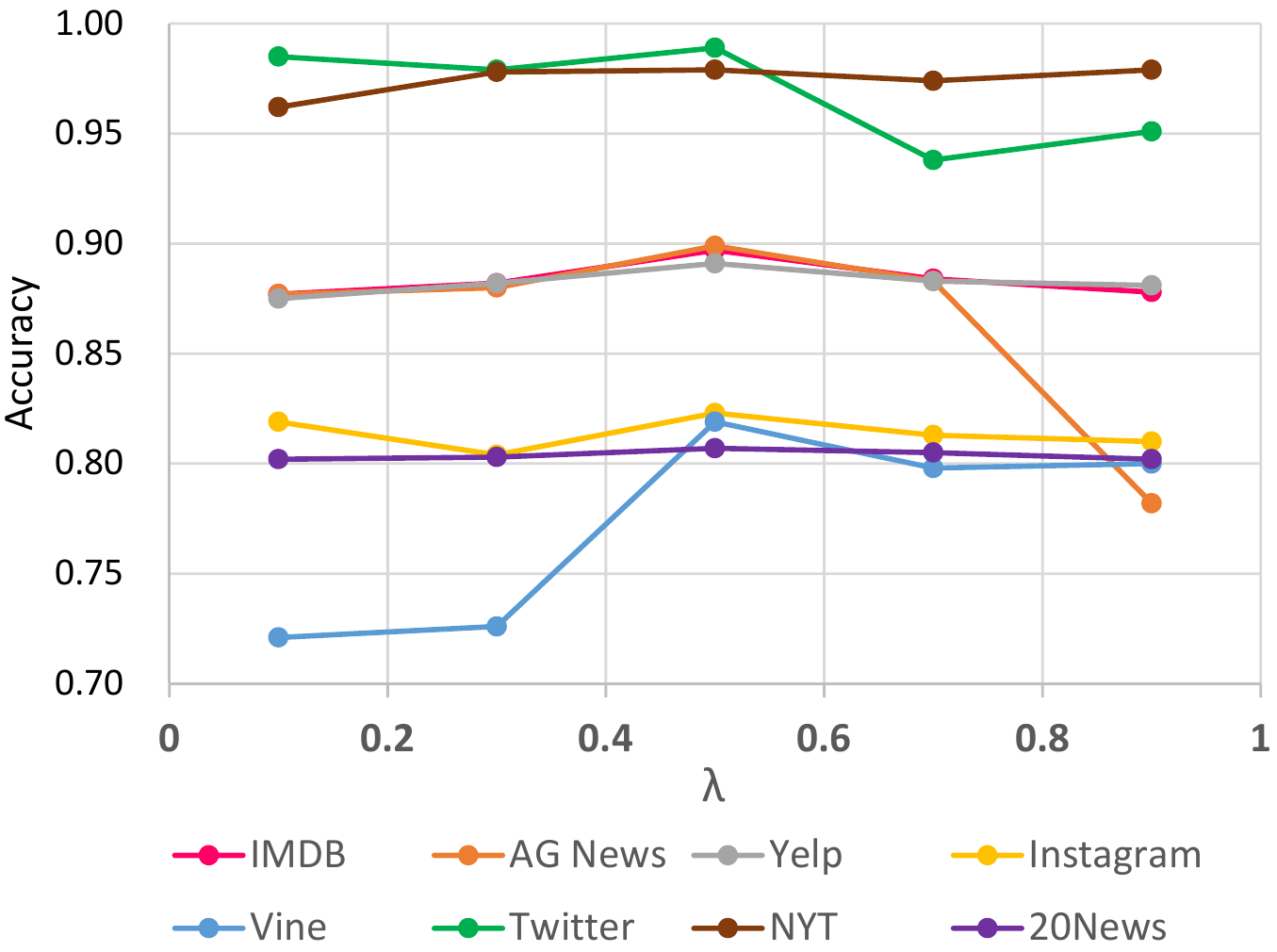}
\caption{Accuracy by varying the factor $\lambda$ in Eq.~\ref{eq-finalobj}.}
\label{exp-lambda}
\end{figure}

\textbf{Balancing Factor $\lambda$.}
We aim at investigating how to weight the loss functions of discretization $J_D$ and classification $J_C$, which is determined by the factor $\lambda$ in Equation Eq.~\ref{eq-finalobj}. By changing the $\lambda$ value to $0.1$, $0.3$, $0.5$, $0.7$, and $0.9$, we show the performance of DDNAS in Figure~\ref{exp-lambda} and answer \textbf{EQ4}. It can be found that $\lambda=0.5$ consistently leads to the highest accuracy scores across datasets. Focusing on minimizing the classification loss (i.e., $\lambda=0.9$) makes the architecture not aware of the potential hierarchical categorization via discretization; hence, the performance is not that good. Paying too much attention to discretization (i.e., $\lambda=0.1$) cannot properly tailor the discovered architecture for text classification, and thus result in an accuracy drop. We would suggest setting $\lambda=0.5$ to balance the two objectives.

\begin{figure*}[!t]
\centering
\includegraphics*[width=1.0\linewidth]{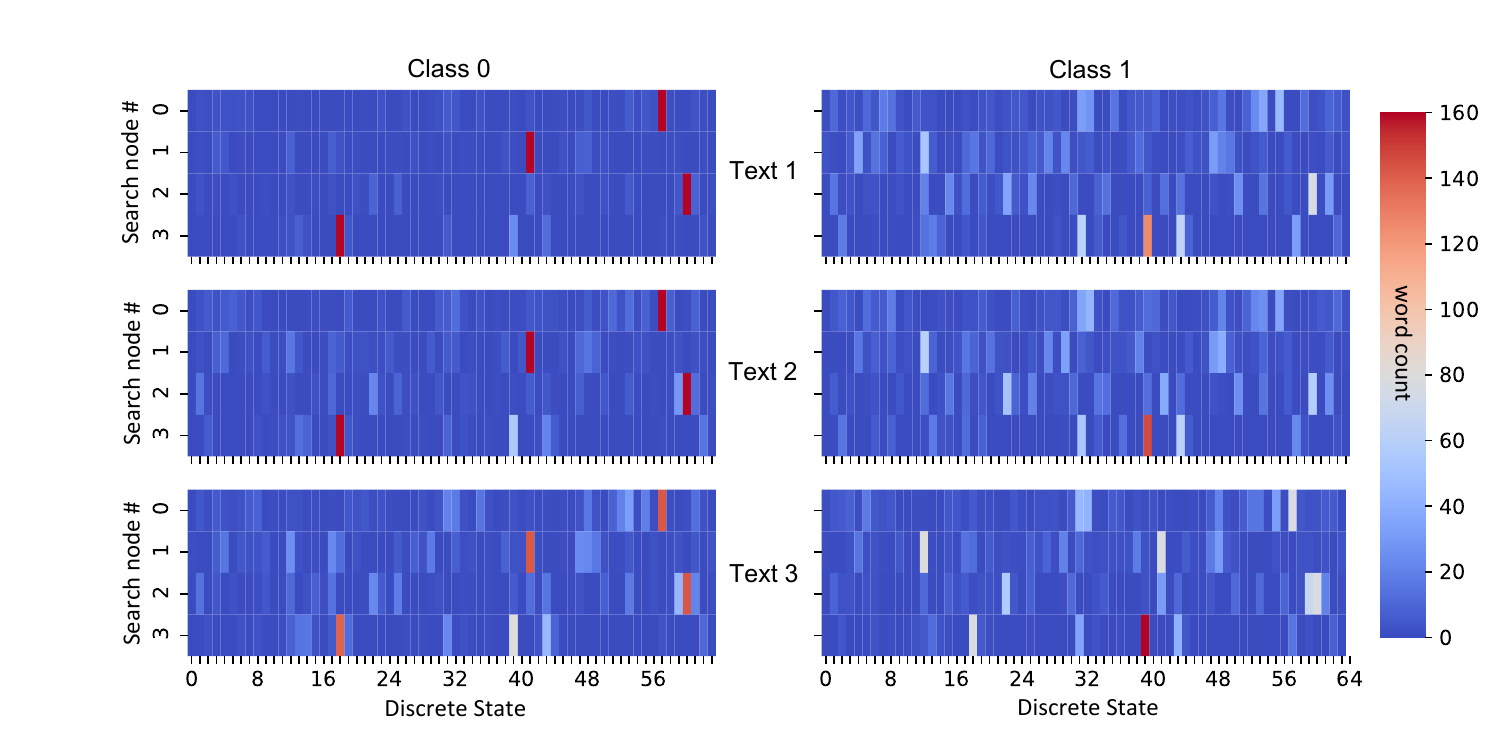}
\caption{Visualization of discretization layers for three testing text documents in IMDB data, in which values in the color bar indicate word count, i.e., the number of words whose representations are discretized into different specific states (one-word representation is discretized into one state).}
\label{exp-visdis}
\end{figure*}

\textbf{Discretization Visualization.}
We answer \textbf{EQ6} by examining whether different classes can be captured by the discretization layer, which is designed to encode latent hierarchical categorization. By randomly selecting six testing instances that belong to two classes in the IMDB dataset, we show their distributions of discretized states using all words in each text instance in Figure~\ref{exp-visdis}, in which the discretization outcome is depicted by the distributions of four search nodes. The distribution displays how many words are discretized into $64$ states across four nodes in the form of a directed acyclic graph, which can be treated as a kind of latent hierarchical categorization. It can be found that instances belonging to different classes exhibit distinct distributions of discretization. Those with the same classes tend to have similar patterns of discretized states. Such results not only demonstrate that the latent hierarchical categorization can be realized via discretization, but also validate its usefulness in text classification.

\begin{figure*}[!t]
\centering
\includegraphics*[width=0.9\linewidth]{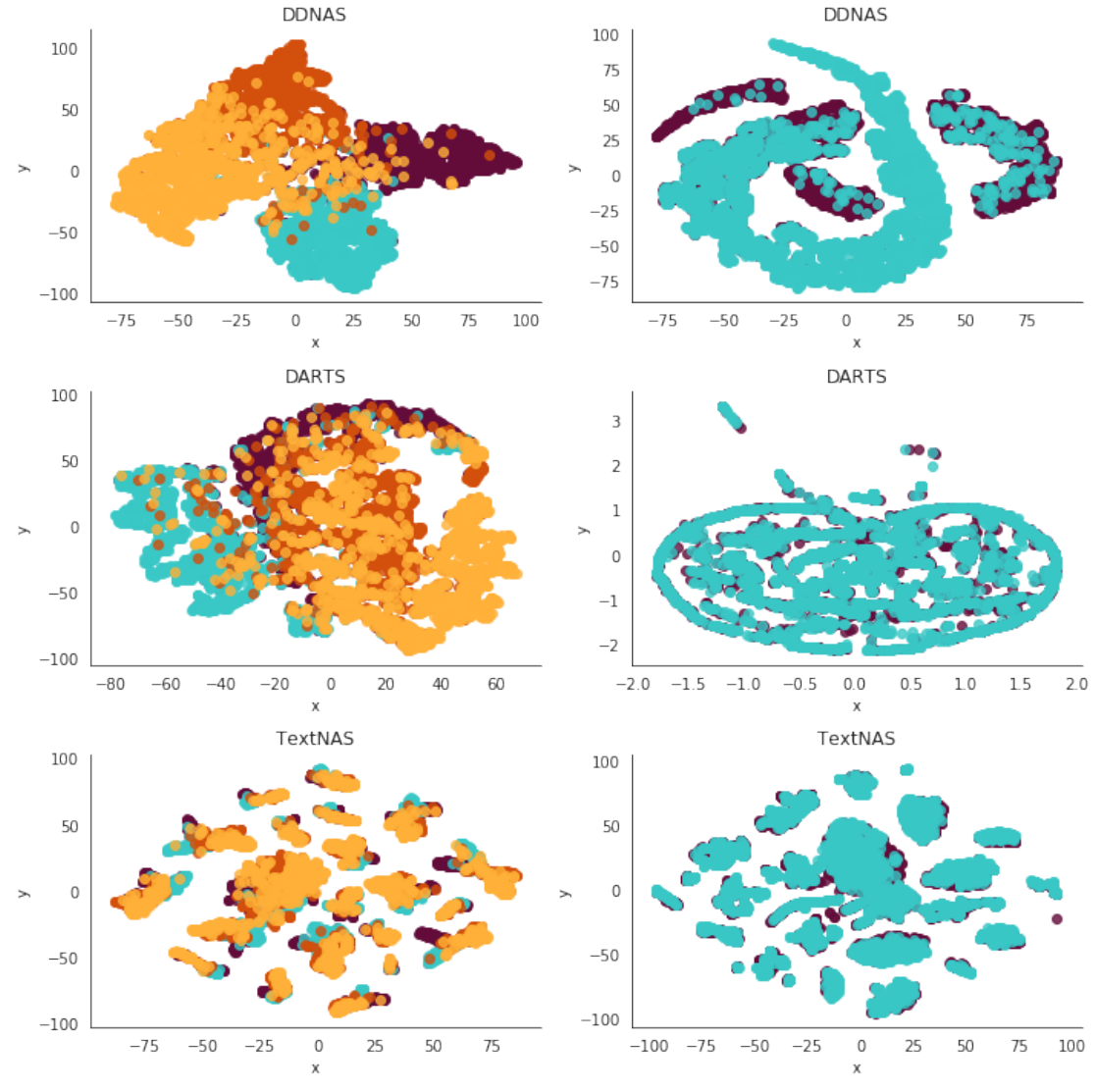}
\caption{t-SNE visualization plots for DDNAS, DARTS, and TextNAS using AG News (left column) and IMBD datasets (right column). Data points with different colors indicate belongingness to different classes.}
\label{exp-tsne}
\end{figure*}

\textbf{Embedding Visualization.} To understand how the model architecture searched by DDNAS outperforms the competitors, we utilize the t-SNE~\cite{tsne2008} visualization tool to produce the plots of embedding visualization for DDNAS, DARTS, and TextNAS using two datasets AG-News and IMBD. The visualization plots are presented in Figure~\ref{exp-tsne}. We can clearly find that the embeddings with different classes generated by DDNAS can better separate from each other on both datasets, compared to DARTS and TextNAS. The model architecture searched by DDNAS with discretized representation learning can push text instances with different classes further away from one another.

\begin{table}[!t]
\centering
\caption{Performance in accuracy on each class of each dataset based on the proposed DDNAS under 20\% training and 80\% testing. The ratio of classes is also reported, where Equal ($n$) means equal ratio for each class and the total number of classes is $n$, and $r_1$:$r_2$ stands for that there are two classes with their corresponding ratios $r_1$ and $r_2$.}
\label{tab:classres}
\resizebox{\columnwidth}{!}{%
\begin{tabular}{c|c|c|c|c}
\hline
 & IMDB & AG News & Yelp & Instagram \\ \hline
Class Distribution & Equal (2) & Equal (4) & Equal (2) & 68\%:32\% \\ \hline
Class Accuracy & 0.834 / 0.850 & 0.78 / 0.82 / 0.81 / 0.71 & 0.883 / 0.852 & 0.855 / 0.772 \\ \hline \hline
 & Vine & Twitter & NYT & 20News \\ \hline 
Class Distribution & 69\%:31\% & 41\%:59\% & Equal (5) & Equal (6) \\ \hline
Class Accuracy & 0.650 / 0.574 & 0.598 / 0.683 & 0.99 / 0.97 / 0.88 / 0.97 / 0.99 & 0.54 / 0.72 / 0.61 / 0.65 / 0.63 / 0.60 \\ \hline
\end{tabular}%
}
\end{table}

\textbf{Class Imbalance \& Error Analysis.}
We aim at investigating how class imbalance affects the performance of DDNAS, and accordingly exploring the potential reasons for the prediction errors. We display the class distribution (i.e., the ratio of each class) and report the accuracy score of each class for the eight datasets based on the proposed DDNAS. The results are shown in Table~\ref{tab:classres}. Among eight datasets, three are a bit imbalanced (i.e., Instagram, Vine, and Twitter) while five are class balanced. For datasets with class imbalance, the minority class tends to have a clearly lower accuracy score, compared to that of the majority class. For datasets whose classes are balanced, their accuracy scores are close to each other. We can obtain two main insights from such results. First, class imbalance does influence the performance of DDNAS. The minority class is relatively challenging to be well trained and predicted. Second, the main classification error comes from testing data instances that belong to the minority class. Given that the data sizes of such three class-imbalanced datasets (i.e., \#Instances) are quite small, as exhibited in Table~\ref{tab-data}, and the number of minority-class instances is very limited, it would be very difficult for DDNAS to have robust model training. That said, a possible reason for classification errors lies in insufficient training data, which prohibits neural network models from producing high testing accuracy.


\begin{table}[!t]
\centering
\caption{Run time across eight datasets and four NAS-based methods under 80\% training and 20\% testing.}
\label{tab:runtime}
\begin{tabular}{c|c|c|c|c}
\hline
 & IMDB & AG News & Yelp & Instagram \\ \hline
ENAS & 4hr & 3hr & 2hr & 20min \\ \hline
DARTS & 2hr & 2hr & 1hr & 10min \\ \hline
TextNAS & 2.5hr & 2.1hr & 1hr & 15min \\ \hline
DDNAS & 2hr & 2hr & 1hr & 12min \\ \hline \hline
 & Vine & Twitter & NYT & 20News \\ \hline
ENAS & 20min & 20min & 4hr & 1hr \\ \hline
DARTS & 9min & 13min & 45min & 30min \\ \hline
TextNAS & 18min & 16min & 1hr & 45min \\ \hline
DDNAS & 12min & 14min & 45min & 30min \\ \hline
\end{tabular}%
\end{table}

\textbf{Time Efficiency.} To understand the time efficiency of the proposed DDNAS, we report the run time for four NAS-based methods, including ENAS, DARTS, TextNAS, and our DDNAS. The execution environment is on GPU machines with NVIDIA Tesla V100 32GB. The results are exhibited in Table~\ref{tab:runtime}. We can find that both DDNAS and DARTS require less run time, compared to ENAS and TextNAS. Given that DDNAS also leads to the best classification performance, as reported in Table~\ref{exp-mainres}, DDNAS is verified to achieve satisfying performance in terms of both effectiveness and efficiency. DDNAS has better time efficiency because it requires neither recurrent cells, nor Transformers as the building blocks, which bring large-scale model parameter size and thus need more training time.

\textbf{Parameter Size.} The parameter size of a model is one of the factors that affect the training efficiency and prediction performance in neural architecture search. Here we report the sizes of model parameters for NAS-based methods. The sizes of model parameters are $4.6$M, $3.3$M, $5$M, and $3.5$M for ENAS, DARTS, TextNAS, and the proposed DDNAS, respectively. It can be obviously found that the parameter size of DDNAS is comparable with DARTS, but is smaller than ENAS and TextNAS. While ENAS and TextNAS contain recurrent cells and Transformers, respectively, as their building blocks, their parameter sizes are much larger. In short, DDNAS is a relatively compact but more effective NAS-based text representation and classification method. And the experimental results also show that it is possible to produce competitive performance without recurrent cells and Transformers in the design of neural architecture for text classification.

\section{Discussion}
\label{sec-discuss}
We discuss various issues regarding the proposed DDNAS, and mainly focus on the flexibility and extensibility, which provide some potential to obtain better performance. That said, DDNAS can serve as a basic NAS framework. We summarize the discussion in the following five points.

\begin{itemize}
\item \textbf{More Building Blocks.} The current DDNAS model considers only three basic operations as the building blocks, i.e., \textit{convolution}, \textit{pooling}, and \textit{none}. We do not incorporate other common operations, such as recurrent layer (e.g., vanilla RNN~\cite{rnn95} and GRU~\cite{gru15}) used by TextNAS~\cite{textnas20} and Zoph and Le~\cite{rlnas17}, and linear transformation with various activation functions including \textit{tanh}, \textit{sigmoid} and \textit{Relu} used by ENAS~\cite{enas18} and DARTS~\cite{darts19}. Existing studies~\cite{textnas20} have proven that the performance of neural architecture search can be benefited by incorporating multi-head self attention~\cite{trm17} and highway connections~\cite{highwaynet}, which are not utilized in DDNAS as well. Given the current DDNAS with only three basic operations can lead to the promising performance of text classification, it is believed that adding a subset of these advanced operations will to some extent improve the effectiveness of DDNAS, and we leave such an attempt as to future work.

\item \textbf{Diverse Tasks.} We have verified the usefulness of DDNAS through the task of text classification. In fact, NAS has widely applied to diverse classification and inference tasks~\cite{nassuv21}, such as image classification~\cite{eanas19} and object detection~\cite{spnas20} in computer vision, node classification and link prediction in social network analysis~\cite{graphnas20}, speech recognition~\cite{dartsasr20} and recommender systems~\cite{autoctr20}. While the latent hierarchical categorization~\cite{hiersuv11} can exist in different types of datasets, DDNAS has a high potential to work effectively in diverse applications. It is also interesting to investigate the correlation between the degree of hierarchical categorization and prediction performance using various tasks and datasets.

\item \textbf{Applications of Discretization Layer.} The proposed discretization layer aims to encode the latent hierarchical categorization by maximizing mutual information. The discretization layer can be imposed into every search node of latent representation under the framework of differentiable architecture search. While DDNAS has been shown to lead to promising performance, it is worthwhile applying the discretization layer to advanced differentiable NAS models, such as stabilizing DARTS~\cite{sdarts20}, progressive DARTS~\cite{pdarts19} and sequential greedy architecture search (SGAS)~\cite{sgas20}, to further boost the performance.

\end{itemize}
%

The novelty of the proposed DDNAS can be highlighted and summarized as the following three main points. First, we propose the discretization layer and impose it into neural architecture search for text representation and classification. The discretization layer can be also seen as a novel kind of regularization in neural architecture search. Second, the macro search space is depicted by a directed acyclic graph (DAG), which is commonly used to represent a class hierarchy in the literatures~\cite{hiersuv11,costa2007review,feng2022hierarchical,HUANG2020105655}. In DDNAS, we impose the discretization layer into each search node in DAG, which allows the NAS learning to capture the latent hierarchical categorization. Both quantitative and qualitative evaluation justify the usefulness of hierarchical categorization. Third, we find that with the discretization layer, the NAS learning for text representation does not require heavy building blocks, such as recurrent cells and Transformers. That said, DDNAS utilizes only operations of convolution, pooling, and none to achieve state-of-the-art performance on the text classification task. DDNAS is experimentally verified to be a simple, compact, effective, and efficient NAS framework for text data. 

\section{Conclusions}
\label{sec-conclude}
In this work, we propose a novel neural architecture search (NAS) algorithm for text classification, Discretized Differentiable NAS (DDNAS). DDNAS is able to jointly discretize the intermediate text embeddings into a discrete space and maximize the classification accuracy, and eventually produce the best-discovered model architecture. The major knowledge learned by the devised discretization layers in the search space of DAG is the latent hierarchical categorization behind the input text. 
Experimental results exhibit that DDNAS can significantly and consistently outperform the state-of-the-art and baselines on eight datasets. Compared to DARTS, the design of discretization layers truly brings an apparent performance boost. In addition, DDNAS can learn effective but compact model architectures from a limited number of training samples, which brings practical usefulness in the context of neural architecture search. In the future, we plan to explore how to impose the discretization layers to other domains' NAS algorithms. We also aim to examine whether DDNAS can also benefit other text-related tasks, such as text generation and summarization, and question answering.

\section{Acknowledgments}
This work is supported by the National Science and Technology Council (NSTC) of Taiwan under grants 110-2221-E-006-136-MY3, 112-2628-E-006-012-MY3, 109-2118-M-006-010-MY2, 111-2221-E-006-001, and 111-2634-F-002-022. This study is also supported by Center for Data Science (CDS), NCKU Miin Wu School of Computing, and Institute of Information Science (IIS), Academia Sinica.

\bibliographystyle{ACM-Reference-Format}
\bibliography{acl2021}


\end{document}